\documentclass{article}

\usepackage{arxiv}

\usepackage[utf8]{inputenc} 
\usepackage[T1]{fontenc}    
\usepackage[pdfauthor={Sajad Darabi, Shayan Fazeli, Jiwei Liu, Alexandre Milesi, Jean-François Puget, Gilberto Titericz}, pdftitle={Heterogenous Ensemble Of Models For Molecular Property Prediction}, pdfkeywords={GNN, Transformer, OGB-LSC, PCQM4Mv2, DFT, Drug Discovery}]{hyperref}
\usepackage{url}            
\usepackage{booktabs}       
\usepackage{amsfonts}       
\usepackage{nicefrac}       
\usepackage{microtype}      
\usepackage{amsmath}
\usepackage{cleveref}       
\usepackage{lipsum}         
\usepackage{graphicx}
\usepackage{natbib}
\usepackage{doi}
\usepackage{siunitx}

\title{Heterogenous Ensemble of Models for Molecular Property Prediction}
\renewcommand{\shorttitle}{Heterogenous Ensemble of of Models for Molecular Property Prediction}
\renewcommand{\headeright}{Technical Report}
\renewcommand{\undertitle}{Technical Report}

\date{}

\author{{\hspace{1mm}Sajad Darabi} \\
	NVIDIA \\
	  USA \\
	\texttt{sdarabi@nvidia.com}\\
\And {\hspace{1mm}Shayan Fazeli} \\
	UCLA \\
	  USA \\
	\texttt{shayan@cs.ucla.edu}\\
\And {\hspace{1mm}Jiwei Liu} \\
	NVIDIA \\
	  USA \\
	\texttt{jiweil@nvidia.com}\\
\And {\hspace{1mm}Alexandre Milesi} \\
	NVIDIA \\
	  France \\
	\texttt{alexandrem@nvidia.com}\\
\And {\hspace{1mm}Pawel Morkisz} \\
	NVIDIA \\
	  USA \\
	\texttt{pmorkisz@nvidia.com}\\
\And {\hspace{1mm}Jean-François Puget} \\
	NVIDIA \\
	  France \\
	\texttt{jpuget@nvidia.com}\\
\And {\hspace{1mm}Gilberto Titericz} \\
	NVIDIA \\
	Brazil \\
	\texttt{gtitericz@nvidia.com}\\
}

\begin{document}
\maketitle

\begin{abstract}
Previous works have demonstrated the importance of considering different modalities on molecules, each of which provide a varied granularity of information for downstream property prediction tasks. Our method combines variants of the recent TransformerM architecture with Transformer, GNN, and ResNet backbone architectures. Models are trained on the 2D data, 3D data, and image modalities of molecular graphs. We ensemble these models with a HuberRegressor. The models are trained on 4 different train/validation splits of the original train + valid datasets. This yields a winning solution to the 2\textsuperscript{nd} edition of the OGB Large-Scale Challenge (2022) on the PCQM4Mv2 molecular property prediction dataset. Our proposed method achieves a test-challenge MAE of $0.0723$ and a validation MAE of $0.07145$. Total inference time for our solution is less than 2 hours. We open-source our code at \href{https://github.com/jfpuget/NVIDIA-PCQM4Mv2}{github.com/jfpuget/NVIDIA-PCQM4Mv2}.
\end{abstract}

\keywords{GNN \and Transformer \and OGB-LSC \and PCQM4Mv2 \and DFT \and Drug Discovery}

\section{Introduction}

The OGB Large-Scale Challenge (LSC) \citep{lsc} is a Machine Learning (ML) challenge to predict a quantum chemical property, the HUMO-LUMO gap of small molecules. This ground truth is obtained via a density-functional theory (DFT) computation which is known to be time-consuming and could take several hours, even for small molecules. With the rapid advancement of machine learning technology, it is promising to use fast, GPU-accelerated and accurate ML models to replace this expensive DFT optimization process.

The PCQM4Mv2 dataset, based on the PubChemQC project \cite{nakata2017pubchemqc}, provides us with a well-defined ML task of predicting the HOMO-LUMO gap of molecules given their 2D molecular graphs. Each molecule has two natural views. The 2D graph incorporates topological structures defined by bonds, and the 3D view provides spatial information that better reflects the geometry and spatial relation of the different bonds in the molecule. In the PCQM4Mv2 dataset, additional 3D structures are also provided for training molecules. Further, in multiple domains, there has been a tendency to convert data modalities to images such that high-capacity and expressive vision models could be used to learn the underlying patterns. Although images do not correspond to the data's natural view, they have shown promising results \cite{kaggle_2021}. We also consider images of the molecule as an alternative view for training vision-based models used in our ensemble. 

Our solution is an ensemble of 39 checkpoints from 10 different models. Before training these models, we performed exploratory data analysis and found a significant distribution shift between the train dataset and the other datasets (valid, test-dev, and test-challenge). The train dataset contains molecules containing at most 20 heavy atoms, while the other 3 datasets contain molecules up to 51 heavy atoms. We made the hypothesis that training on the proposed train split would lead to mediocre performance on large molecules (more than 20 atoms).

We decided to use different train/validation splits by creating $24$ folds from the combined train+valid datasets. We then trained a model on $23$ folds and used the last fold for validation. Ideally, one would train 24 models that way, using each fold as a validation fold in turn. This is the usual $k$-folds cross-validation, with $k=24$. However, for practical running time considerations, we decided to only use 4 folds as validation folds, \textit{i.e.} training 4 fold models. Each model is trained by selecting one of the first 4 folds as validation fold, and using the remaining 23 folds as training data. Whenever we discuss folds in the remainder of this report we refer to the aforementioned 4 validation folds.

In the following sections of this report, we describe the model variants trained for each data modality, followed by a a section describing how these models were ensembled for the final test-challenge prediction.

\section{Transformer-M variants}
\label{transformerm}

 Following prior work incorporating 2D and 3D information via bias terms in the Transformer self-attention \citep{ying2021do}, we use the recently proposed random structural channels between 2D/3D in Transformer-M \citep{2022transformerm} as our backbone architecture. In this method, the same Graphormer \citep{ying2021do} backbone architecture is used for both channels.

Given the input molecule $\mathcal{G}_{\text{2D/3D}}$, our encoder outputs a latent representation $z = f(\mathcal{G_{\text{2D/3D}}})$. We leverage the provided 3D SDF file to extract the corresponding 2D \& 3D spatial atom positions for this model. In subsequent subsections, we describe the variants used.

All hyperparameters not mentioned in the following sections (dropout, gradient clipping, weight decay, learning rate schedule, number of kernels) were kept the same as described in the Transformer-M work.

\subsection{$\textrm{Transformer-M}^\textrm{base}_\textrm{without\_denoising}$}
The first variant is a 12-layers, 768-wide model that includes randomly-activated structural channels with probabilities $\frac{1}{4}$ for 2D, $\frac{1}{2}$ for 3D, and $\frac{1}{4}$ for 2D+3D. In this variant, the denoising task was not included, but the input atom positions provided by the organizers were perturbed with gaussian noise of magnitude 0.2 in the 3D channel. This variant was trained once for each of the 4 folds, for 400 epochs with batch size 1024, on an NVIDIA DGX-1 (8 V100-SXM2-16GB 160W).

\subsection{$\textrm{Transformer-M}^\textrm{large}_\textrm{with\_denoising}$}

The second variant is similar to the first one, except for its 18 layers, and the inclusion of the atom position denoising task, using a cosine similarity loss weighted equally to the graph prediction loss. Stochastic depth was enabled with a probability of 10\%. This variant was trained once for each of the 4 folds, for 400 epochs with batch size 1024, on an NVIDIA DGX-2H (16 V100-SXM3-32GB 300W).

\subsection{$\textrm{Transformer-M}^{\textrm{large}}_{\textrm{baseline}}$}
In this variant, we train the Transformer-M model without any modifications. The random channels probabilities are set to the original weights $\frac{1}{5}$, $\frac{1}{5}$, $\frac{3}{5}$ for 2D, 3D, and 2D+3D respectively. The training setup is the same as previous variants, except the training lasts 454 epochs. This variant was trained twice with different seeds on the 4 folds.

\subsection{$\textrm{Transformer-M}^{\textrm{large}}_{\textrm{Dirichlet}}$}
In this variant, we set the random channels probabilities to follow a Dirichlet distribution with weights $\frac{1}{5}$, $\frac{1}{5}$, $\frac{3}{5}$ for 2D, 3D, and 2D+3D respectively. The training setup is the same as previous variants, except the training lasts 454 epochs. This variant was trained once on the 4 folds.

\subsection{Transformer-M with Knowledge Guided Regularization} 
Research in the domain of molecular property prediction has shown that incorporating additional knowledge-based information could lead to considerable performance improvements. 

Specifically, the observations in \cite{li2022kpgt} show that pre-training a model on molecule fingerprints and descriptors configured as a mask-and-predict pretext task can result in significant improvements, and with merely relying on 2D information even surpassing most of the works leveraging 3D positions.

Given that PCQM4Mv2 is a large-scale dataset and considerably larger than those considered in \cite{li2022kpgt}, we formulated a similar task as a {\it regularizing} objective, built on top of the Transformer-M configuration for predicting the HOMO-LUMO gaps.
For each molecule in the dataset, we followed the configuration in \cite{li2022kpgt} and computed a descriptor $\mathbf{y}_{\text{d}} \in \mathbb{R}^{200}$ and a fingerprint $\mathbf{y}_{\text{fp}} \in \{0,1\}^{512}$.
We added two additional projection heads on top of the latent graph representations, and added the following regularization loss:

\begin{equation}
    \mathcal{L}_{\text{reg}}=\lambda_{\text{fp}} \mathcal{L}_{\text{bce}}(g_{\text{fp}}(\mathbf{z}),\mathbf{y}_{\text{fp}})
    + \lambda_{\text{d}} \mathcal{L}_{\text{mse}}(g_{\text{d}}(\mathbf{z}),\mathbf{y}_{\text{fp}})
\end{equation}

\begin{figure}[!h]
    \centering
    \includegraphics[scale=0.35]{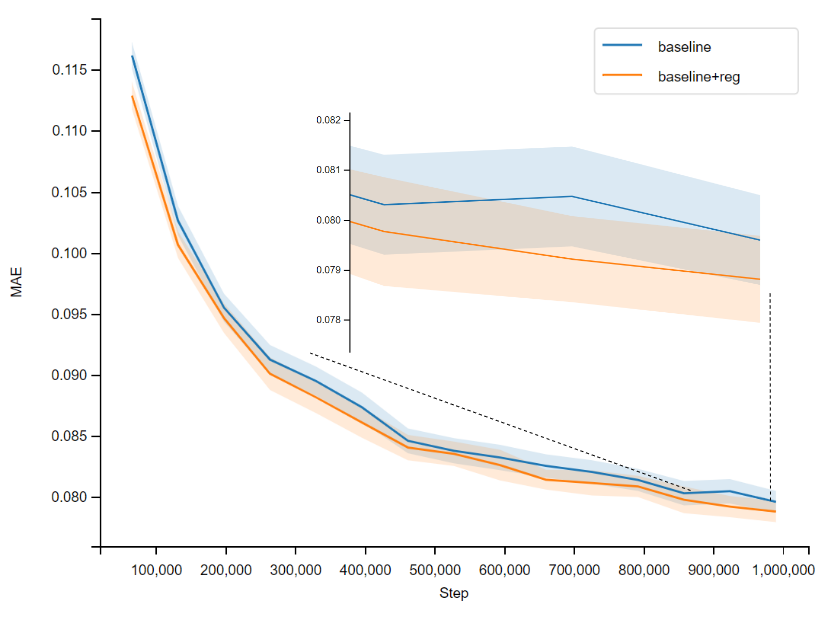}
    \caption{Comparison of Transformer-M (baseline) and Transformer-M with knowledge-guided regularization (baseline+reg)}
    \label{fig:tmkpgt}
\end{figure}

We run initial series of experiments using the base architecture with the number of layers set to 12, batch size 1024, learning rate $\num{2e-4}$ and end learning rate $\num{1e-9}$, with exponential decay over the learning rate schedule for 300 epochs. We fix the random seed for both experiments to the same value, and as shown from Figure \ref{fig:tmkpgt} this addition shows an improvement of $0.007$ MAE over the baseline.

We train two models with the added regularization on the 4 folds with varying weights, one set to $\lambda=0.1$ and another set to $\lambda=0.2$. The hyperparameters are set to the same values as the large model with 18 layers. We call these models $\textrm{Transformer-M}_{\textrm{kpgt}_{\lambda=0.1}}^{\textrm{large}}$ and  $\textrm{Transformer-M}_{\textrm{kpgt}_{\lambda=0.2}}^{\textrm{large}}$,  respectively. The models are trained on 8 NVIDIA V100-SXM2-16GB 160W GPUs.

\section{Molecular Transformer}
\label{molecular transformer}

For each input molecule, we build a graph using the available 2D graph information from SMILES. This graph is built using data from an instance of the \texttt{ogb} \texttt{PCQM4Mv2Dataset} class. We extend the \texttt{smiles2graph} function by adding rings using the \texttt{rdkit} package:
\begin{verbatim}
        ssr = Chem.GetSymmSSSR(mol)
        ssr = [list(sr) for sr in ssr]
\end{verbatim}
We use this extended \texttt{smiles2graph} function to preprocess the dataset.

In the graph for a given molecule, every node is initialized with a token embedding of size 768. Embeddings are initialized from a Glorot uniform distribution. The nodes are:
\begin{itemize}
    \item A node for the whole molecule. Its role is similar to the \texttt{[VNode]} in Graphormer \citep{ying2021do}. This token is initialized with an embedding of the total number of heavy atoms in the molecule, capped at 20. In addition to that, the inverse of the number of atoms is added to the embedding.
    \item A node for each heavy atom in the molecule. The atom token is initialized with the \texttt{ogb} predefined embedding plus an embedding of the degree of the atom in heavy atom graph of the molecule.
    \item A node for each bond in the molecule. Its token is initialized with the \texttt{ogb} predefined embedding.
    \item A node for each ring returned by \texttt{RDKit} \texttt{GetSymmSSSR()}. Each ring token is initialized with an embedding of the number of atoms in the ring, capped at 15.
\end{itemize}

The edges of the graph are constructed as follows:
\begin{itemize}
    \item For each bond between atom $A$ and atom $B$, create an edge between the node for $A$, and the node for $B$.
    \item For each bond between atom $A$ and atom $B$ create an edge between $A$ and $C$, and create an edge between $B$ and $C$, where $C$ is the node for the bond.
    \item For each ring $R$, and for each atom $A$ in the ring, create an edge between the node for $R$ and the node for $A$.
    \item An edge between the node for the whole molecule and each of the other nodes.
\end{itemize}

Tokens are fed into a 12-layers Transformer \citep{transformer}. Changes compared to vanilla Transformers are:
\begin{itemize}
    \item Attention is masked by the edges defined above: a node only attends its neighbors in the graph. We found that it was more efficient to compute dense attention then mask rather than use sparse attention, probably because the graph is quite small and dense.
    \item The vanilla transformer feed-forward layer after attention layer is replaced by two vanilla transformer feedforward layers.
    \item Following the recent trends, we use GELU activation \citep{gelu}.
\end{itemize}

Attention is masked by subtracting 10 from vanilla attention when there is no edge, before softmax. There is no other attention mask. We tried to use the Graphormer attention bias (spatial embedding, etc.), but it did not improve our model. We think that the explicit representation of bonds and rings captures the topology of the graph as effectively. Our model performance on the original validation dataset is indeed slightly better than that of Graphormer.

The final layer is a dense layer taking as input the last embedding for the molecule node.

Our model is trained with a sine annealing learning rate scheduler, \textit{i.e.}, starting with a very low learning rate, up to a maximum and down again following a sine curve. The maximum learning rate was set at $\num{2e-4}$, and the minimum at $\num{1e-8}$. We trained for 30 epochs and a batch size of 256 on 4 NVIDIA V100 16GB 160W GPUs using \texttt{PyTorch DataParallel}. We experimented with \texttt{DistributedDataParallel}, but we could not achieve the same model accuracy for some reason.
\section{PD-DGN: Pointwise Dense Pooling for Deep Graph Neural Networks}
\label{PD-DGN}

Equivariant GNN models are widely used for the task of graph prediction tasks, we design a novel pointwise dense pooling instead of the conventional global mean pooling for deep graph neural networks (PD-DGN) model. The idea is similar to self-attention graph pooling (SAG) \cite{lee2019self} but simpler. SAG considers both node and topology features while our design only utilizes node features. We train the model with 4 different folds and average the predictions. The approach achieved a $0.089$ mean absolute error on the validation split.

\subsection{Model Architecture}
\begin{figure}[!h]
	\centering
	\includegraphics[scale=0.38]{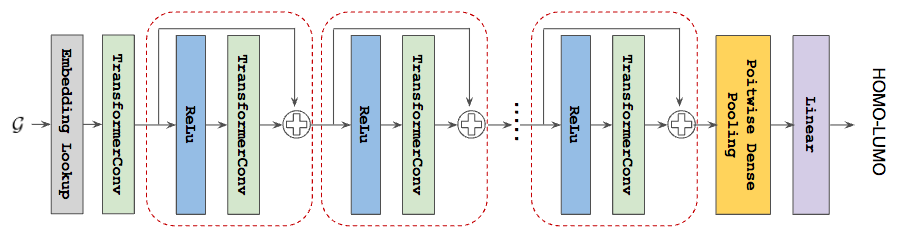}
	\caption{Architecture of our deep graph Transformer. The boxes with red dashed lines are \texttt{DeepGCNLayer} with skip connections. }
	\label{fig:net}
\end{figure}

Similar to the Molecular Transformer variant in section \ref{molecular transformer}, this variant uses the given \texttt{PygPCQM4Mv2Dataset} class of the \texttt{ogb} package, which converts SMILES strings to graphs. Each node and each edge is represented by a vector of dimensions of 9 and 3, respectively. We use embeddings to encode the node features. The overall architecture of our model is shown in Figure~\ref{fig:net}. In this model the core building block is \texttt{DeepGCNLayer} \cite{li2019deepgcns} which is a wrapper of \texttt{TransformerConv} \cite{shi2020masked} with additional skip connections. This building block \texttt{DeepGCNLayer} enables the net to grow deeper, mitigating the over-smoothing problem commonly seen as GNNs become deeper. We use a total of 15 \texttt{DeepGCNLayer} blocks in our network.

The \texttt{TransformerConv} implements the following operations,
\begin{equation}
\mathbf{x}^{\prime}_i = \mathbf{W}_1 \mathbf{x}_i +
        \sum_{j \in \mathcal{N}(i)} \alpha_{i,j} \mathbf{W}_2 \mathbf{x}_{j}
\end{equation}
where $\mathbf{x}_i$ and $\mathbf{x}^{\prime}_i$ are the input and output embeddings of node $i$ of a \texttt{TransformerConv} layer, respectively. $\mathcal{N}(i)$ represents the neighbors of node $i$. The attention coefficients $\alpha_{i,j}$ are computed via multi-head dot product attention:
\begin{equation}
\alpha_{i,j} = \textrm{softmax} \left(
        \frac{(\mathbf{W}_3\mathbf{x}_i)^{\top} (\mathbf{W}_4\mathbf{x}_j)}
        {\sqrt{d}} \right)
\end{equation}
where $d$ stands for the dimension of the embedding vectors $x_i$ and $x_j$.
At the end of the last \texttt{TransformerConv}, we have transformed embeddings for each node. To convert node embeddings to graph embeddings $\mathbf{G}$, global mean pooling is often used, as shown in equation~\ref{eq:mean}. 

\begin{equation}
\mathbf{G} = \frac{1}{N} \sum_{i=1}^{N} \mathbf{x}_i.
\label{eq:mean}
\end{equation}

With global mean pooling, each node contributes equally to the graph embedding, which could be suboptimal. An intuitive idea to improve global mean pooling is to use a weighted sum of the node embeddings to compute graph embedding, where weights are learned from node embeddings. We propose a novel pointwise dense pooling layer:

\begin{equation}
\mathbf{G} = \frac{1}{N} \sum_{i=1}^{N} \alpha_{i} \mathbf{x}_i.
\label{eq:pd}
\end{equation}

where $\alpha_{i}$ is computed via $\mathbf{x}_i$ using a learnable parameter $\mathbf{W}$:

\begin{equation}
\alpha_{i} = \textrm{softmax} \left(
        \mathbf{W} \mathbf{x}_i
        \right)
\end{equation}

\subsection{Experiment Results}
Our model is implemented using PyTorch Geometric framework \cite{Fey/Lenssen/2019} and we train on an NVIDIA DGX-1 (8 V100-SXM2-16GB 160W). The model was trained with 4 different folds of the concatenated data of the given training and validation dataset. Training one model with one seed on a single GPU takes 10 hours. Inference with one model and one seed on the test-dev split takes 50 seconds. Experimental results showed that the proposed pooling has at least 0.001 lower MAE for both validation and test-dev than the baseline global pooling. 


\section{Convolutional Neural Network Model}
\label{cnns}
CNNs are well known to be very efficient in solving a large range of image challenges. In this chapter, we describe a CNN regression approach to predict the HOMO-LUMO energy gap of molecules using image-based representation.

\subsection{ResNet Backbone}
\begin{figure}[!h]
	\centering
	\includegraphics[scale=0.38]{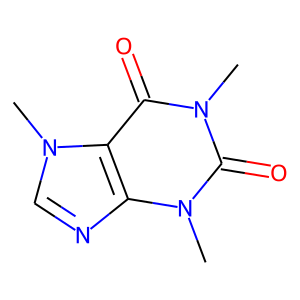}
	\caption{2D image molecule representation of caffeine}
	\label{fig:rdkitmol}
\end{figure}

The model uses a regular ResNet34 \citep{resnet} image backbone with the top classification head replaced by a linear layer with one output. The images are created using \texttt{RDKit MolToFile} function that takes as input the \texttt{Mol} instance created from the smiles representation. The image representation of the smiles \texttt{Cn1c(=O)c2c(ncn2C)n(C)c1=O} (caffeine) can be seen in Figure \ref{fig:rdkitmol}. To speed up the training, the model training has three phases that use different images sizes and learning rates combinations. During the phases, the images are resized starting from 224x224, then 288x288 and finally to 352x352 pixels. The learning rate scheduler used was cosine annealing with restarts. Also, image augmentations that helped to regularize better include horizontal and vertical flips, shift, scale, and rotation. Augmentations start hard in the initial phase of training, and for the final phase, only horizontal flips are applied to fine-tune the model. This model achieves an MAE around $0.098$ on the validation set. The loss function used was plain L1.

\subsection{EfficientNet Backbone}

Using an EfficientNet-B3 \citep{efficientnet} backbone achieves an MAE of $0.092$ on the validation set, which outperforms ResNet34 backbone, but this model didn't improve the final ensemble metric. So we didn't use it for the final solution.


\section{Stacking \&  Results}
\label{stacking}

We have described how we trained a set of heterogeneous models. We are now going to describe how we ensembled this set of models. For the sake of clarity, we will refer to the models described so far as base models. 

\subsection{Base Model Results}

In table \ref{tab:results} we summarize the fold MAE results for each base model. Several runs are missing (denoted with $-$) due to training failures and lack of time to restart training for final submission, hence we exclude their MAE values in this table.
\begin{table*}[h]%
    \renewcommand{\arraystretch}{1.6}
    \caption{\label{tab:results} Model mean absolute error (MAE) results on 4 folds.}
    \begin{minipage}{\textwidth}
    \begin{center}
    \begin{tabular}{l | ccccc}
    \toprule
    & \multicolumn{4}{c}{\textbf{MAE}} \\
    \textbf{Model} & \textbf{Fold 0} & \textbf{Fold 1} & \textbf{Fold 2} & \textbf{Fold 3} \\
    \hline
    $\textrm{Transformer-M}^{\textrm{large}}_{\textrm{baseline}_{seed=3407}}$ & {-} & {0.07424} & {0.07416} & {0.07414}\\
    $\textrm{Transformer-M}_{\textrm{kpgt}_{\lambda=0.2}}^{\textrm{large}}$ & {0.07477} & {-} & {0.07417} & {0.07410}\\
    $\textrm{Transformer-M}_{\textrm{kpgt}_{\lambda=0.1}}^{\textrm{large}}$ & {0.07478} & {0.07448} & {-} & {0.07421}\\
    $\textrm{Transformer-M}^{\textrm{large}}_{\textrm{baseline}_{seed=42}}$ & {0.07469} & {0.07434} & {0.07416} & {-}\\
    $\textrm{Transformer-M}^{\textrm{large}}_{\textrm{Dirichlet}_{seed=3407}}$ & {0.07434} & {0.07409} & {0.07386} & {0.07396}\\
    $\textrm{Transformer-M}^\textrm{base}_\textrm{without\_denoising}$ & {0.07645} & {0.07583} & {0.07591} & {0.07596}\\
    $\textrm{Transformer-M}^\textrm{large}_\textrm{with\_denoising}$ & {0.07865} & {0.07744} & {0.07892} & {0.07926}\\
    Molecular Transformer & {0.079534} & {0.079366} & {0.079198} & {0.079090} \\
    PD-DGN & {0.084606} & {0.084109} & {0.084103} & {0.084054} \\
    CNN EfficientNet-B3 & {0.092380} & {0.092456} & {0.092663} & {0.092523} \\
    CNN ResNet34 & {0.098702} & {0.097201} & {0.099301} & {0.098005} \\
    \hline
    \end{tabular}
    \end{center}
    \end{minipage}
\end{table*}

\subsection{Stacking}

The simplest way to ensemble our base models is to average their output. This works fine when the models have similar accuracy.  However, when we ensemble heterogeneous base models with a wide range of accuracy, we want to favor the best base models over weaker models. One way to achieve this is to use a weighted average of the base models' outputs. Strong base models should get a high weight and weaker base models a small weight. This requires computing effective weights. 

A general purpose for computing these weights is to train a linear model with a new training dataset. The input in this training dataset is made of the predictions of the base models, and the target output is the original target.  This is often referred to as stacking, because the linear model sits on top of the base models' predictions.

In order to train the linear model we need to have predictions of the base models on data for which we also have ground truth. This is why we created 4 train/validation folds as described in the introduction.
We follow the following workflow, for each base model and each fold:
\begin{itemize}
    \item We train a fold base model. 
    \item We store the predictions of the fold base model on its validation fold. These predictions are called out-of-fold (OOF) predictions.
    \item We store the predictions of the fold base model on the test-challenge data
\end{itemize}

In some of our training runs, one of the folds failed (a diverging loss during training), hence we only had 3 valid checkpoints for the base model. This happened for 4 of the runs. We decided to replace the out of fold predictions for these 4 base models with their average where they ran correctly. We applied the same averaging for the test predictions.

Then for each base model, we concatenate its 4 out-of-fold predictions. We then concatenate the result for all base models in the column dimension. This yields a new dataframe with as many columns as base models and as many rows as the sum of rows of the 4 validation folds we used.  This is the training data for the linear model.

The ground truth for training the ensemble model is the target of the 4 validation folds we use.

We then tried several linear models for ensembling. The best result was obtained using \texttt{scikit-learn} \texttt{HuberRegressor}. \texttt{HuberRegressor} was chosen because it performed best to optimize the L1 loss metric when compared to other training algorithms, \textit{e.g.}, Linear Regression, Lasso, Random Forest, and Gradient Boosting Decision Trees.

We used 4 folds cross validation to train the linear model, using the same 4 validation folds used for the base models. A linear model is trained using 3 of the folds, and evaluated on the last fold. We then compute the average score across 4 validation folds. 

Running 4-fold cross-validation metric scored \textbf{0.07145} overall for the linear models.

To generate test predictions, a similar process is applied:
\begin{itemize}
    \item The base fold models are grouped by the validation fold used to train them.
    \item For each validation fold, compute test predictions of all fold base models for that fold
    \item For each validation fold, apply the linear model trained with that validation fold to the test predictions of the fold models. This yields a test predictions
    \item Average the 4 linear models test predictions.
\end{itemize}

This yields a first ensemble of models.  The weights of fold models computed by the Huber regressor are given in Table \ref{tab:weights}.  The first row is the average of predictions for these base models: $\textrm{Transformer-M}^{\textrm{large}}_{\textrm{baseline}_{\textrm{seed}=3407}}$, $\textrm{Transformer-M}_{\textrm{kpgt}_{\lambda=0.2}}^{\textrm{large}}$, $\textrm{Transformer-M}_{\textrm{kpgt}_{\lambda=0.1}}^{\textrm{large}}$, and 
$\textrm{Transformer-M}^{\textrm{large}}_{\textrm{baseline}_{\textrm{seed}=42}}$. The last row is the Huber regressor intercept.

\begin{table*}[h]%
    \renewcommand{\arraystretch}{1.6}
    \caption{\label{tab:weights} Model weights from Huber regressor.}
    \begin{minipage}{\textwidth}
    \begin{center}
    \begin{tabular}{l | ccccc}
    \toprule
    & \multicolumn{4}{c}{\textbf{weights}} \\
    \textbf{Model} & \textbf{Fold 0} & \textbf{Fold 1} & \textbf{Fold 2} & \textbf{Fold 3} \\
    \hline
    $\textrm{Transformer-M}$ average & {0.5382} & {0.5330} & {0.5186} & {0.5171}\\
    $\textrm{Transformer-M}^{\textrm{large}}_{\textrm{Dirichlet}_{\textrm{seed}=3407}}$ & {0.2356} & {0.2447} & {0.2564} & {0.2657}\\
    $\textrm{Transformer-M}^\textrm{base}_\textrm{without\_denoising}$ & {0.1013} & {0.0986} & {0.0936} & {0.0945}\\
    $\textrm{Transformer-M}^\textrm{large}_\textrm{with\_denoising}$ & {-0.0353} & {-0.0413} & {-0.0269} & {-0.0349}\\
    Molecular Transformer & {0.0929} & {0.0974} & {0.0878} & {0.0878} \\
    PD-DGN & {0.0395} & {0.0394} & {0.0418} & {0.0421} \\
    CNN ResNet34 & {0.0255} & {0.0261} & {0.0265} & {0.0254} \\
    intercept & {0.0101} & {0.0096} & {0.00991} & {0.0094} \\
    \hline
    \end{tabular}
    \end{center}
    \end{minipage}
\end{table*}

Near the end of the competition, we retrained some of the base models on the full train+valid data, hoping that these full train base models would be different and slightly better than the fold base models. These models were added to the ensemble predictions via a simple weighted average according the following equation. 

$$
\begin{aligned}
\textrm{submission} = (\textrm{ensemble} &+ 0.2552 \cdot \textrm{Transformer-M}^{\textrm{large}^{\textrm{full\_train}}}_{\textrm{dirichlet}}\\ &+ 0.1747 \cdot \textrm{Transformer-M}_{\textrm{kpgt}_{\lambda=0.1}}^{\textrm{large}^\textrm{full\_train}}\\ &+ 0.1747 \cdot \textrm{Transformer-M}_{\textrm{kpgt}_{\lambda=0.2}}^{\textrm{large}^\textrm{full\_train}}) / (1+0.2552+0.1747+0.1747)
\end{aligned}
$$

The weight 0.2552 is the average weight for $\textrm{Transformer-M}^{\textrm{large}}_{\textrm{Dirichlet}_{\textrm{seed}=3407}}$ in Table \ref{tab:weights}. The weight 0.1747 is $1/3$ of the average weight for $\textrm{Transformer-M}$ average in Table \ref{tab:weights}.

\subsection{Inference Time}

\begin{table*}[h]%
    \renewcommand{\arraystretch}{1.6}
    \caption{\label{tab:timing} Inference and dataset timing for each model.}
    \begin{minipage}{\textwidth}
    \begin{center}
    \begin{tabular}{l | cc}
    \toprule
    \textbf{Model} & \textbf{Dataset Time (s)} & \textbf{Inference (s)} \\
    \hline
    $\textrm{Transformer-M}^{\textrm{large}}_{\textrm{baseline}_{seed=3407}}$ & {150} & {45}\\
    $\textrm{Transformer-M}_{\textrm{kpgt}_{\lambda=0.2}}^{\textrm{large}}$ & {150} & {45}\\
    $\textrm{Transformer-M}_{\textrm{kpgt}_{\lambda=0.1}}^{\textrm{large}}$ & {150} & {45}\\
    $\textrm{Transformer-M}^{\textrm{large}}_{\textrm{baseline}_{seed=42}}$ & {150} & {45}\\
    $\textrm{Transformer-M}^{\textrm{large}}_{\textrm{Dirichlet}_{seed=3407}}$ & {150} & {45}\\
    $\textrm{Transformer-M}^\textrm{base}_\textrm{without\_denoising}$ & {150} & {30} \\
    $\textrm{Transformer-M}^\textrm{large}_\textrm{with\_denoising}$ & {150} & {30} \\
    Molecular Transformer & {300} & {70}  \\
    PD-DGN & {60} & {60} \\
    CNN EfficientNet-B3 & {120} & {319}  \\
    CNN ResNet34 & {120} & {312} \\
    \hline
    \end{tabular}
    \end{center}
    \end{minipage}
\end{table*}
Inferencing is pretty efficient. If we ignore dataset download from OGB site then Transformers take less than a minute on average to make test predictions using a single V100 GPU. Making all Transformers predictions takes about 30 minutes. The Resnet model takes a bit longer, about 5 minutes per fold model. In total, inferencing takes about 1 hour for all base models. Running the ensemble model on top of their predictions takes a minute or so.

Most of the time is spent on the dataset download and smiles2graph or smiles to image conversions. We use 4 different variants of the dataset when inferencing: 
\begin{itemize}
    \item A vanilla \texttt{PCQM4Mv2Dataset} for all $\textrm{Transformer-M}$ variants \ref{transformerm}.
    \item A \texttt{PCQM4Mv2Dataset} extended with ring information from rdkit for the Molecular Transformer base model \ref{molecular transformer}.
    \item A \texttt{PygPCQM4Mv2Dataset} for the PD-GDGN base model \ref{PD-DGN}.
    \item A \texttt{PCQM4Mv2Dataset} extended with images generated by \texttt{rdkit} for the ResNet34 base model \ref{cnns}.
\end{itemize}

Each dataset variant is constructed only once and shared by all corresponding base models.

We created these datasets on the same hardware used for training, i.e. with a multicore CPU. We measured the time taken to create these datasets. We then estimated the time taken to build the dataset for the test challenge prediction to be 4 percent of the total time to build the dataset (refer to Table \ref{tab:timing} for dataset \& inference timings). Estimated running times for the test challenge part range from a minute for $\textrm{Transformer-M}$ variants to 5 minutes for ResNet34 and for the Molecular Transformer. 

Taking all times into account we estimate our inference time to be around 1h30, and certainly less than 2 hours.

\section{Conclusion}
 We present a novel winning solution based on an ensemble of heterogeneous models. Our best models are variants of the recently published Transformer-M architecture. We get a significant improvement by ensembling these variants with weaker models, namely other Transformers, and an image classification model leveraging additional 2D/image data modalities. Our solution demonstrates the effectiveness of ensembling heterogeneous models on large-scale datasets and we hope this will encourage others to pursue such avenues more often in similar problems.
 
\bibliographystyle{unsrtnat}
\bibliography{references}
\end{document}